\documentclass[10pt,twocolumn,letterpaper]{article}

\usepackage{3dv}
\usepackage{times}
\usepackage{epsfig}
\usepackage{graphicx}
\usepackage{amsmath}
\usepackage{amssymb}

\usepackage{color}
\usepackage[skip=5pt]{caption}
\usepackage{dsfont}
\usepackage{subfig}
\usepackage{multirow}
\usepackage{booktabs}
\usepackage{authblk}

\usepackage[pagebackref=true,breaklinks=true,letterpaper=true,colorlinks,bookmarks=false]{hyperref}

\threedvfinalcopy 

\def\threedvPaperID{16} 

\ifthreedvfinal\fi
\begin{document}

\newcommand{\R}{\mathbb{R}}
\newcommand{\itemset}{\cal{I}}
\newcommand{\itm}{i}
\newcommand{\itmtwo}{j}
\newcommand{\costterm}{E}
\newcommand{\numsamples}{N_s}
\newcommand{\cparams}{\Phi}
\newcommand{\cparamsnew}{\Phi^{*}}
\newcommand{\contextfunc}{\pi}
\newcommand{\scene}{S}
\newcommand{\argmin}{\operatornamewithlimits{argmin}}
\newcommand{\argmax}{\operatornamewithlimits{argmax}}


\title{Pano2CAD: Room Layout From A Single Panorama Image}

\author{
Jiu Xu$^1$ $\qquad$ Bj\"orn Stenger$^1$ $\qquad$ Tommi Kerola$^*$ $\qquad$  Tony Tung$^2$\thanks{The work was done while the authors were with Rakuten.} \\
$^1$ Rakuten Institute of Technology $\quad$
$^2$ Facebook
}

\maketitle

\begin{abstract}
This paper presents a method of estimating the geometry of a room and the 3D pose of objects from a single 360$^{\circ}$ panorama image. Assuming Manhattan World geometry, we formulate the task as a Bayesian inference problem in which we estimate positions and orientations of walls and objects. The method combines surface normal estimation, 2D object detection and 3D object pose estimation.
Quantitative results are presented on a dataset of synthetically generated 3D rooms containing objects, as well as on a subset of hand-labeled images from the public SUN360 dataset.
\end{abstract}

\section{Introduction}
3D scene understanding from images has been an active research topic in computer vision, enabling applications in navigation, interaction, and robotics. 
State-of-the-art techniques allow layout estimation from a single image of an indoor scene~\cite{DelPeroCVPR2013,SatkinIJCV2014,SchwingECCV2012}, which is an underconstrained problem.
Most prior work estimates the layout of a room corner only or assumes a simple box-shaped geometry.
Since a standard camera lens has a limited field of view, an incremental procedure is usually necessary to recover a whole scene~\cite{CabralCVPR2014}.
A simple alternative is to capture panorama images of the scene, assuming that objects of interest are visible.
For example, the {\it PanoContext} method~\cite{ZhangECCV2014} recovers the full room layout from one panorama image, while still assuming a box-shaped room. Walls and floor are used as context information to recognize object categories and positions.

\begin{figure}[t]
\begin{center}
\includegraphics[width=0.99\linewidth]{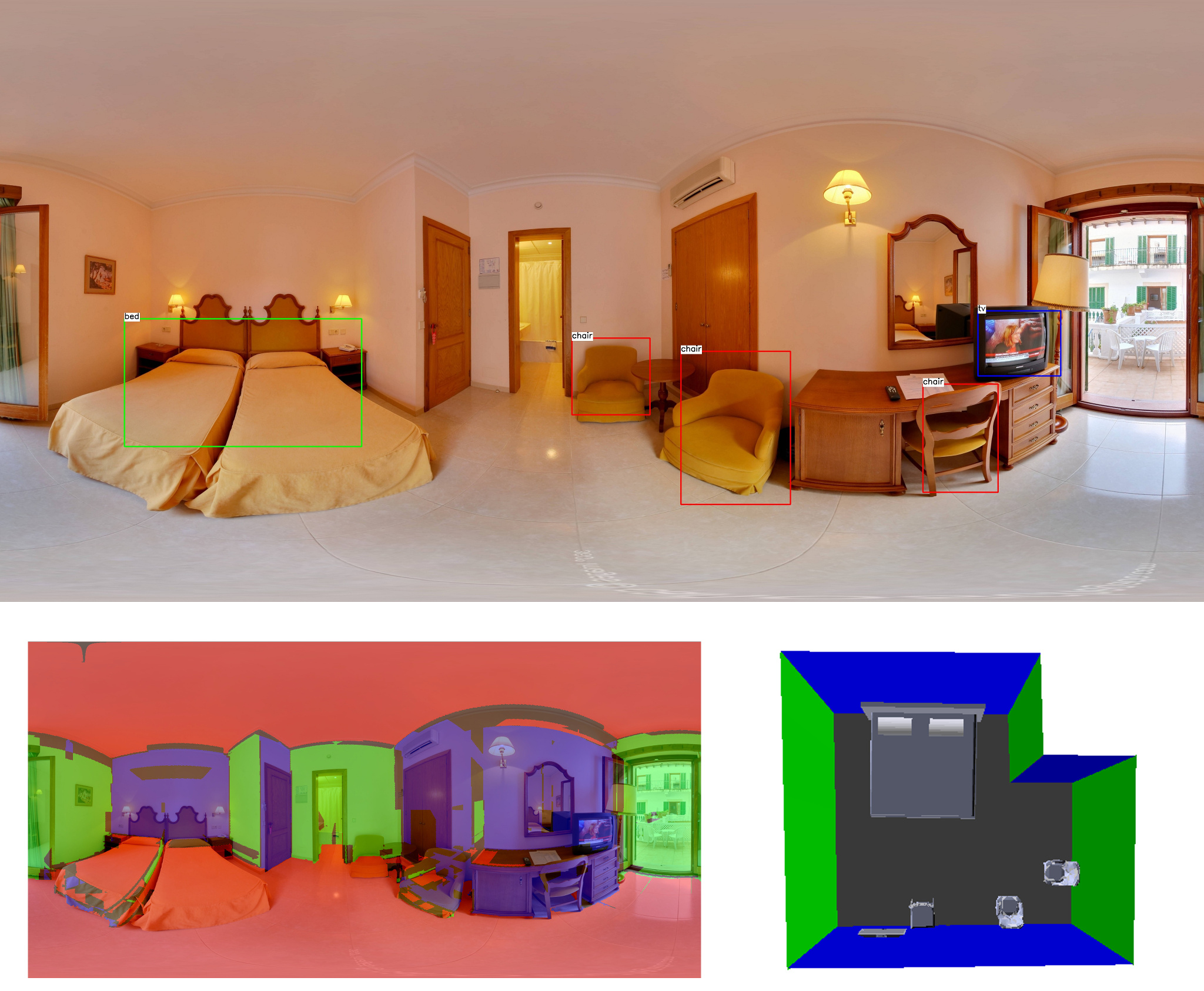}
\vspace*{-1em}
\end{center}
   \caption{\textbf{Example output:} \it Indoor scene reconstruction from a single panorama image. Top: input image with 2D object detection bounding boxes. Detection is carried out in perspective images and the bounding box coordinates are projected into the panorama image. Bottom left: estimated surface orientations. Bottom right: reconstructed 3D room geometry and furniture items (top view).}
\label{fig:teaser}
\end{figure}

In this paper, we build on insights from the {\it PanoContext} work, but no longer assume a rectangular floor plan.
In contrast to the bottom-up object proposals from edges~\cite{ZhangECCV2014}, we employ more robust top-down methods for object detection and 3D pose estimation.
To accomplish this, we first transform the single panorama image into a set of perspective images from which we estimate per-pixel surface orientations and object detections. From these we obtain a first scene layout up to an unknown scale.
Next, objects are detected using a trained detector and initial 3D poses are estimated using a libary of 3D models. Global scale is estimated indirectly by projecting 3D object models of known dimensions into the scene. We sample room hypotheses and evaluate their posterior probability. 
See Fig.~\ref{fig:teaser} for an example output of our algorithm.
%
%


The contributions of this work are:
(1) We relax the box-shape assumption of~\cite{ZhangECCV2014} to a Manhattan World assumption, reconstructing the complete shape of the room.
(2) Object location and pose is estimated using top-down object detection and 3D pose estimation using a public library of 3D models,
(3) We introduce a context prior for object and wall relationships in order to sample plausible room hypotheses.
We evaluate the accuracy of the method on synthetically generated data of 3D rooms as well as results on images from the public SUN360 dataset. Please also see the supplementary video for qualitative results.

\section{Related work}

We put our work into context by discussing prior work in the areas of surface estimation from images, 3D object models, and context priors.

\paragraph{Geometry estimation. }
Early seminal work in layout estimation includes surface estimation from an outdoor image~\cite{HoiemIJCV2007} by learning a mapping from an input image to a coarse geometric description.
Similarly, the {\em Make3D} method estimates a 3D planar patch model of the image, where the training data consists of images and corresponding depth maps~\cite{SaxenaPAMI2009}.
More recently much progress has been made estimating pixel-wise normals from images~\cite{EigenICCV2015, HaeneCVPR2015}.
For indoor scenes, assuming a Manhattan World geometry, vanishing points can be detected and the camera parameters recovered. 
For example, Lee \etal~\cite{LeeCVPR2009} proposed a method to interpret a set of line segments to recover 3D indoor structure, demonstrating that the full image appearance is not necessary to solve this problem.
Hedau \etal~\cite{HedauICCV2009} modeled the whole room as a 3D box and learned to classify walls, floor, ceiling, and other objects in a room.
Work by Schwing \etal~\cite{SchwingCVPR2012,SchwingECCV2012}  estimates a 3D box-shaped room from a single image using integral geometry for efficiently evaluating 3D hypotheses.
Earlier work by Yu \etal~\cite{YuCVPR2008} takes a bottom-up edge grouping approach to infer set of depth-ordered planes.
The work by Wang \etal~\cite{WangACM2013} has shown improved accuracy by estimating cluttered areas, including all objects except the room boundaries.
In this paper we estimate surface orientations of the whole scene~\cite{HedauICCV2009,LeeCVPR2009} and treat the orientations in object regions separately.
Other approaches include Cabral and Furukawa~\cite{CabralCVPR2014}, who use multiple input images to apply 3D reconstruction and estimate a piece-wise planar 3D model.
Building on the work by Ramalingam and Brand~\cite{RamalingamICCV2013}, recent work by Yang and Zhang~\cite{YangCVPR2016} recovers 3D shape from lines and superpixels in a constraint graph.
Complementary to these two methods we estimate the 3D room geometry together with 3D objects inside. 


%
%
%
\paragraph{3D Objects.}

Objects contained within a room have been modeled at different levels of complexity.
For example, Lee~\etal~\cite{LeeNIPS2010} fit 3D cuboid models to image data, demonstrating that including volumetric reasoning improves the estimation of the room geometry.
Hedau \etal \cite{HedauECCV2010,HedauCVPR2012} showed that the scene around an object is useful for building good object detectors, however it was also limited to cuboid objects like large pieces of furniture.
Del Pero \etal~\cite{DelPeroCVPR2013} proposed part-based 3D object models, allowing more accurate modeling of fine structures, such as table legs. Configurations of their detailed models are searched using MCMC sampling.
In their 'Box in the Box' paper, Schwing \etal~\cite{SchwingICCV2013} used a branch-and-bound method to jointly infer 3D room layout and objects that are aligned with the dominant orientations.
Satkin \etal~\cite{SatkinIJCV2014} proposed a top-down matching approach to align 3D models from a database with an image. The method employs multiple features to match 3D models to images, including pixel-wise object probability, estimated surface normals, and image edges.
In recent work by Su \etal, a CNN was trained for pose estimation for 12 object categories (from the PASCAL 3D+ dataset) from rendered 3D models~\cite{SuICCV2015,xiangWACV2014}.
Tulsiani \etal~\cite{TulsianiPAMI2016} combine object localization and reconstruction from a single image using CNNs for detection and segmentation, and view point estimation. This top-down information is fused with shading cues from the image.
While these are viable approaches, the number of categories is limited and our object shapes of interest are typically not represented exactly. 
We therefore estimate 3D object models from a model database, similar to the recent work in~\cite{HuangSIGGRAPH2015}.

\paragraph{Context priors. }

Pieces of furniture tend not to be uniformly distributed within a room but follow certain constraints that include physical constraints, such as non-intersection, or somewhat less rigid functional constraints, such as aligning a bed with one of the walls or leaving some space to access all areas of the room.
Such prior knowledge has been employed to improve layout estimation.
For example, Del Pero \etal~\cite{DelPeroCVPR2012,DelPeroCVPR2013} introduced constraints to avoid object overlap and to explicitly search for objects that frequently co-occur, such as tables and chairs.
In {\emph PanoContext}, Zhang \etal~\cite{ZhangECCV2014} show that context evidence of an entire room can be captured from panoramic images. They learn pairwise object displacements to score their bottom-up object hypotheses. However, their box-shaped room model does not take relative orientation or distance to walls into account.
Some insight can be gained from the graphics literature where generative models have been used for 3D model search.
For example, Fisher and Hanrahan proposed a method for efficient  search of 3D scenes~\cite{FisherTOG2010}. Pairwise relationships were learned from scene graphs from 3D Warehouse, but only take relative distances, not orientation, into account.
Merrel \etal~\cite{MerrellSIGGRAPH2011} proposed a density function for room layout design that encodes numerous design rules, such as respecting clearance distance around objects and the relative alignment of objects with each other.
Handa \etal~\cite{HandaArXiv2015} used geometric constraints to automatically generate 3D indoor scenes as training data.
The method proposed here scores room layout hypotheses, in a 2D top-down view, with pairwise energy terms, encoding object-to-object and object-to-wall constraints, but allows for more flexibility compared to the generative model in~\cite{MerrellSIGGRAPH2011}.

\begin{figure*}[t!]
\begin{center}
 \includegraphics[width=0.95\linewidth]{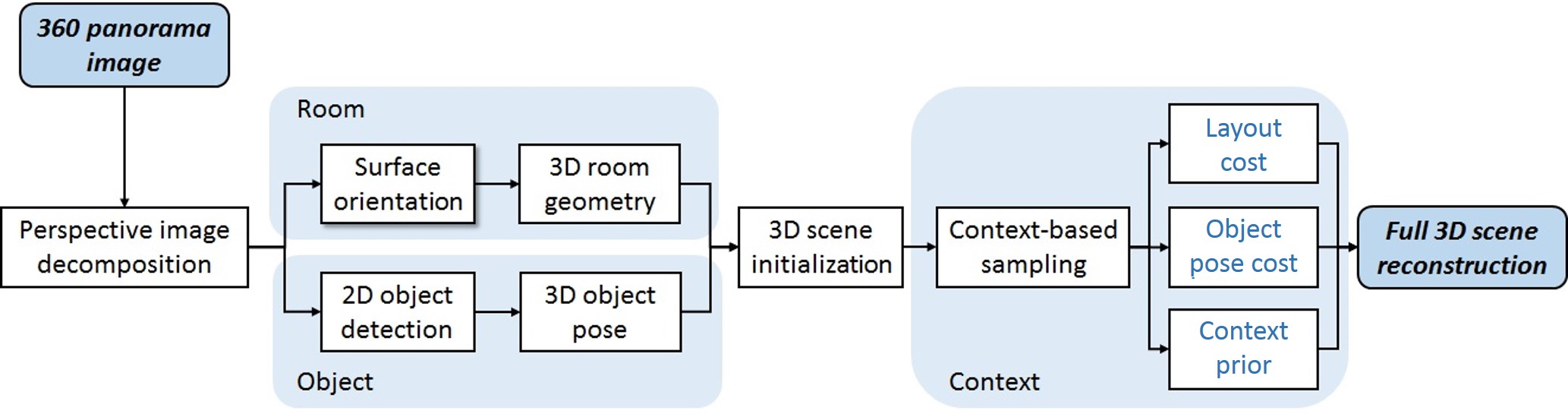}
\end{center}
   \caption{\textbf{Algorithm overview.} \it From a single panorama image the proposed method estimates initial 3D room geometry and 3D object poses. Subsequently we sample a global posterior distribution which includes terms for room layout, object poses, and a context prior.}
\label{Fig2_flowchart}
\end{figure*}

\section{Generative model}

Given an indoor scene $\scene=(W, O)$, defined by a set of walls $W=\{w_i\}_{i=1}^{N_w}$ and a set of objects $O=\{o_j\}_{j=1}^{N_o}$,
we formulate the room layout estimation in a Bayesian framework, where the model  parameters consist of
\begin{equation}
 \Phi = (c, \lambda,  p^w_i, \theta^w_i,   p^o_j, \theta^o_j  ), 
\end{equation}
 which includes the camera model $c$, the absolute room scale $\lambda$, wall center positions and wall orientations $\{p^w_i, \theta^w_i\}$, as well as center positions $\{p^o_j\}$ and the orientations $\{\theta^o_j\}$ of objects  $\{o_j\}$ in the scene $\scene$.
We formulate the estimation task as maximizing the probability $P(\Phi|\mathcal{I})$
of the model parameters $\Phi$ given an input image $\mathcal{I}$ of the scene $\scene$.
This is equivalent to maximizing the posterior  $P(\mathcal{I}|\Phi)\pi(\Phi)/P(\mathcal{I})$ to obtain 
the set of optimal parameters:
\begin{equation}
\label{eqn:map}
\Phi_{\mathrm{MAP}}  = \argmax_{\Phi} \, P(\mathcal{I}|\Phi) \, \pi(\Phi) \, ,
\end{equation}
where $\pi(\Phi)$ is the prior on the model parameters,  and $P(\mathcal{I})$ is assumed uniform.
In what follows, we give details on the different components of the model and the estimation process. The overall approach is summarized in Fig.~\ref{Fig2_flowchart}.


\begin{figure}[t!]
\centering
\subfloat[]{\includegraphics[width=0.9\columnwidth]{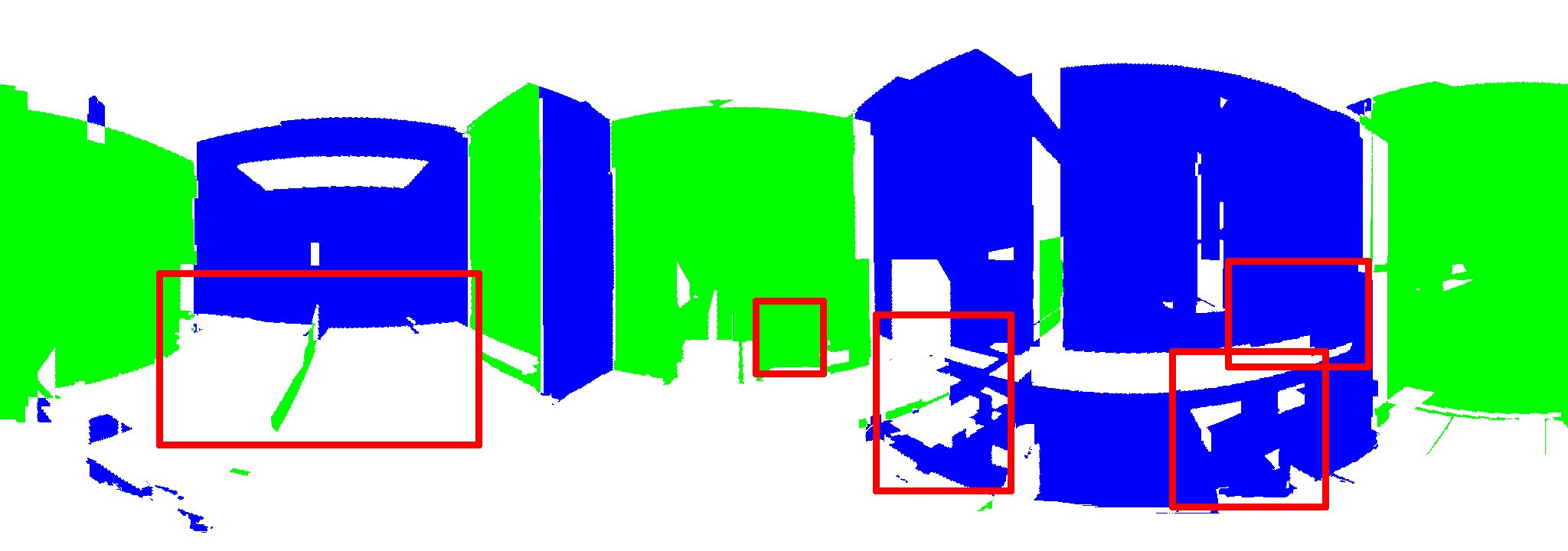}} \\
\subfloat[]{\includegraphics[width=0.9\columnwidth]{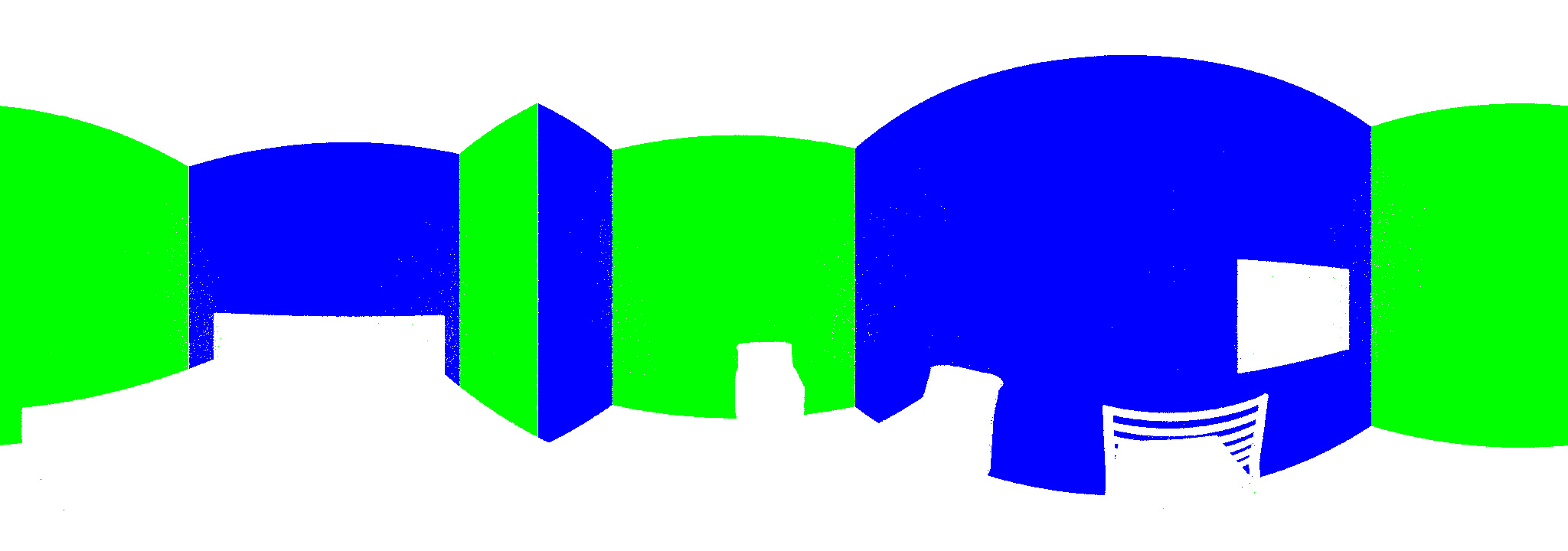}} \\
\subfloat[]{\includegraphics[width=0.45\columnwidth]{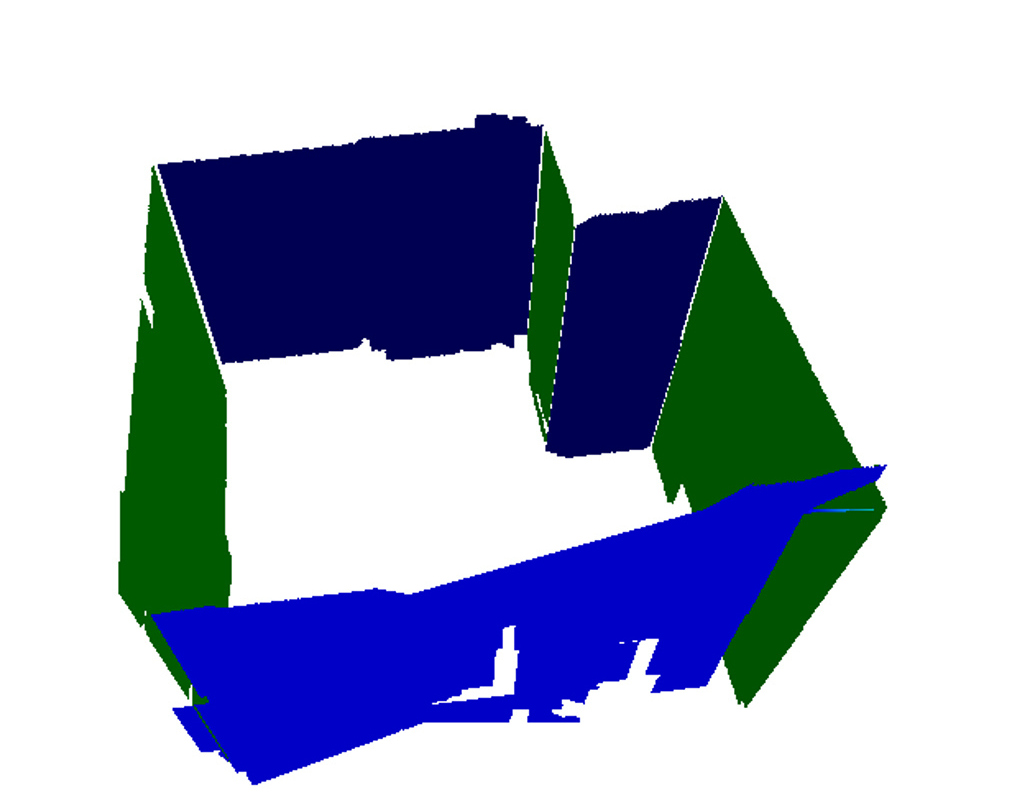}} \hspace*{1em}
\subfloat[]{\includegraphics[width=0.45\columnwidth]{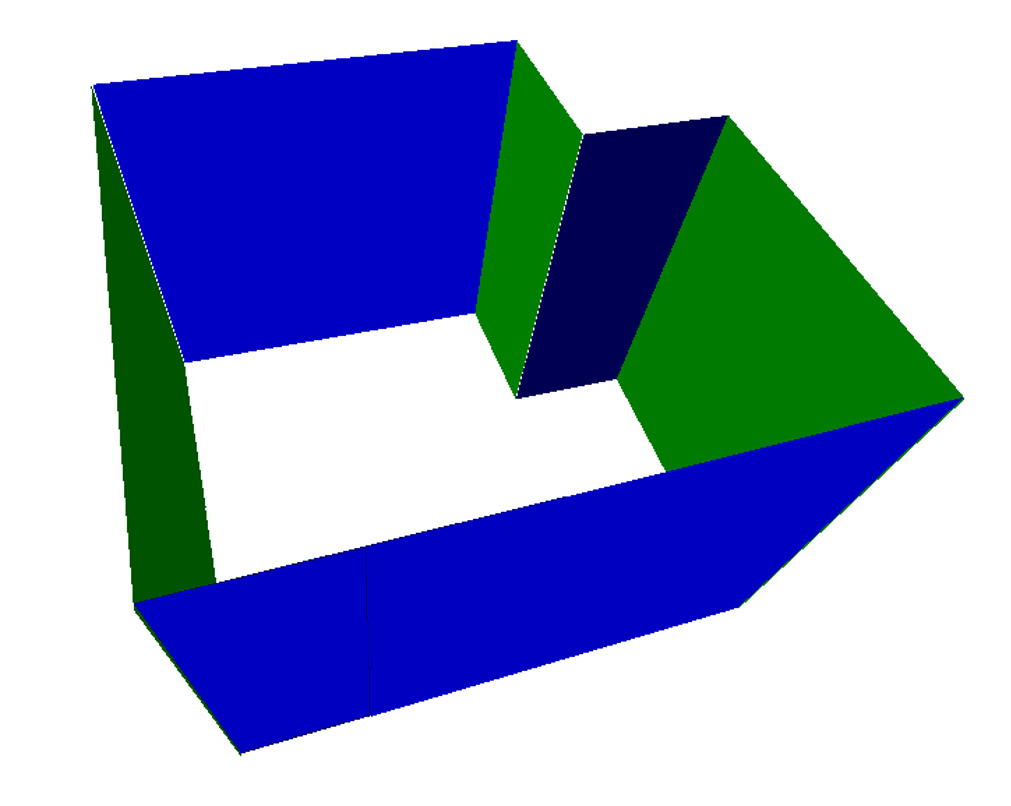}}
\vspace*{1em}
\caption{\textbf{Processing steps:} \it (a) Surface orientations of the observed input image with detected object regions used for masking. (b) Surface orientations of rendered image of predicted models with occluding objects (\ie, silhouettes serve as masks).   (c) Room geometry after surface alignment at unknown scale. (d) Room geometry after plane fitting.}
\label{Fig3_OM2Room}
\end{figure}

\subsection{Room layout likelihood}

We decompose the likelihood in Eq.~\ref{eqn:map} as follows, making a conditional independence assumption:
\begin{equation}
\label{eqn:lhood}
P(\mathcal{I}|\Phi)  = P(\mathcal{I}|\lambda, p^o_j, p^w_i, \theta^w_i) \, P(\mathcal{I}| \theta^o_j ) ,
\end{equation}
obtaining two likelihood terms, one for the scene including object positions, and one for object orientations. Both terms are evaluated by comparing the projected image $\mathcal{D}$ of the predicted 3D scene model (\ie, obtained with estimated model parameters), with the observed orientation image. 
Our justification for decomposing the likelihood is that we synthesize only the walls of the scene, not the (occluding) objects. Object bounding boxes, which are not always accurate, serve as masks. Therefore we do not account for object orientation when evaluating the first term.
Note that like in previous work the camera parameters $c$ are approximated by placing it at the center of the spherical image at a known height, \eg 1.70m for SUN360 dataset~\cite{ZhangECCV2014}. 

In the following we describe the processing steps to evaluate the first likelihood term. We first transform the spherical panorama input image $\mathcal{I}$  into a set of $K$ perspective images 
$\{\mathcal{I}_{k}\}_{k=1}^{K}$, which no longer contain strong distortions.
This transformation returns a set of perspective images with overlapping regions, in our case 6 images with a 90$^{\circ}$ field of view and 30$^{\circ}$ of overlap between adjacent images.
For each image $\mathcal{I}_{k}$ we estimate surface orientations at each pixel by combining estimates of their Orientation Map (OM)~\cite{LeeCVPR2009} and Geometric Context (GC)~\cite{HedauICCV2009}. We apply GC to the panorama and combine the OM and GC in the floor region to obtain wall positions and orientations~\cite{ZhangECCV2014}.
We segment the panorama image into regions of three orthogonal surface normal directions (see Fig.~\ref{fig:teaser}, bottom left).
The orientation surface image for each $\mathcal{I}_{k}$  is used to recover partial 3D room geometry.
The surface normals in each perspective image are converted to 3D points using vanishing points and camera-to-floor distance using the method in~\cite{DelageISRR2005}.
Each image corresponds to a separate 3D point cloud with unknown scale. We apply the constraint that corresponding pixels in overlapping regions in the images have the same depth:
We globally align the point clounds by minimizing the sum of 3D point distances of points corresponding to the overlapping image regions. This is followed by greedy plane fitting, starting from the largest segment, using the Iterative Closest Point algorithm~\cite{Besl1992}, resulting in an initial estimate of 3D room geometry, \ie positions and orientations of walls  $\{\hat{w}_i\}$, up to scale, see Fig.~\ref{Fig3_OM2Room}.

We also run object detection in each image $\{\mathcal{I}_{k}\}_{k=1}^{K}$ using a {\it Faster R-CNN} (details in Sect.~\ref{sec:experiments}). The coordinates of object locations are reprojected to the panorama image $\mathcal{I}$, and non-maximum suppression is applied to eliminate redundant detections.
Since the camera $c$ is oriented toward the center of $\mathcal{I}$, assuming the center position at 0$^{\circ}$, the positions $\{p_j\}$ of detected objects  $\{o_j\}$ can be derived from the polar coordinates. Absolute distances of objects to the camera remain unknown at this stage.
Having estimated a 3D scene model including walls and object positions, we define the likelihood term for the room layout as
\begin{equation}
P(\mathcal{I}| \lambda, \{p^o_j\}, \{p^w_i,\theta^w_i \})  \propto \exp \left[ -E_s(\mathcal{I}, \{p^o_j\}, \{p^w_i,\theta^w_i \} )\right].
\end{equation}
The cost function $E_s$ evaluates a room hypothesis by reprojecting the synthesized 3D scene back into the panoramic view and compare surface normals:
\begin{align}
    & E_s(\mathcal{I}, \{p^o_j\}, \{p^w_i,\theta^w_i \} ) = 1 - \frac{N_{\mathrm{c}} }{N_{\mathrm{pix}}} \; , \quad \mathrm{where} \\
  & N_{\mathrm{c}} = \sum_{m\in {\mathcal I}} \mathds{1}_{ l(\mathcal{I}_m) =  l(\mathcal{D}_m)} (m) \; .
\end{align}
The cost is low when surface orientations of the predicted 3D scene agrees with the orientation image $\mathcal{I}$.
The terms $l(\mathcal{I}_m)$ and $l(\mathcal{D}_m)$  are the discrete surface orientation labels at pixel location $m$ in the surface orientation maps of images $\mathcal{I}$ and $\mathcal{D}$, respectively, $\mathds{1}$ is the indicator function, and $N_{\mathrm{pix}}$ is the number of pixels in $\mathcal{I}$.
In previous work~\cite{DelPeroCVPR2012,LeeCVPR2009}, a similar term is used to evaluate wall geometry hypotheses. However, the presence of objects and occluded walls in the scene, which add noise to the estimation, were not considered.
Here, we propose to mask detected objects in the surface orientation images as follows:
In the observed image, bounding boxes of detected objects serve as masks, while in the predicted images, silhouettes of 3D objects serve as masks, see Fig.~\ref{Fig3_OM2Room} (a) and (b). Hence $E_s$ is evaluated in image regions of visible wall regions.
Since visible wall areas are directly related to object size, the pixel-wise cost function $E_s$ is sensitive to the global scale $\lambda$ and object positions $\{p_j\}$.
For example, if the estimated room scale $\lambda$ is smaller than the true scale, objects in the synthetic scene will be placed closer to the camera, thereby occluding larger wall regions, which is penalized by the cost function $E_s$.



\subsection{Object pose estimation} 
\label{sec:orientation}
We define the second factor in the likelihood term in Eq.~\ref{eqn:lhood} as:
\begin{equation}
P(\mathcal{I}|\{\theta^o_j\})   \propto \exp \left[-E_o(\mathcal{I}, \{\theta^o_j\}) \right],
\end{equation}
where the cost function $E_o$ evaluates object orientation hypotheses $\{\theta^o_j\}$ by comparing HOG descriptors~\cite{dalal2005histograms} of detected objects in $\mathcal{I}$ and rendered images from corresponding 3D models.
For the initial object pose estimation $\{\hat{\theta}_j\}$ (superscript omitted for clarity in this section), two distinct sources of data are employed: a set of rendered images $\mathcal{R}$ of 3D models with known pose, and a set of visually similar web images $\mathcal{W}$ found by Google Image search.
The auxiliary set of images helps to regularize the solution when jointly estimating object pose, as demonstrated in~\cite{HuangSIGGRAPH2015} and confirmed in initial experiments.
We therefore take the same approach as~\cite{HuangSIGGRAPH2015} with two extensions. First, we do not assume images with clean background and therefore extract the object in the input image by automatic grab-cut segmentation, assuming that the image center contains the object and image corners are part of the background.
Further, [14] assumes that a very specific category type is known (e.g. Windsor chair). Our approach works with just knowing the abstract category (chair), and uses visual search to find web images similar to the target object. Therefore our approach is more robust against cluttered background and generalizes to a wider range of object categories. The web images are obtained automatically by retrieving the first 400 results of Google Image search for visually similar images within the detected object category. Background is removed from the web images by co-segmentation, since we assume that this image set contains a shared common object~\cite{dong2015interactive}.

Given object bounding boxes, HOG descriptors are computed for each region in a $4 \times 4$ image-grid using unsigned gradients with  $\ell_2$-normalization, and are concatenated into a global image descriptor. A CRF model is then employed to regularize the pose estimation.
Let $\mathcal{T}$ denote the cropped input image (or multiple images if the same object appears more than once in the scene) of objects for which we want to find the 3D pose $\theta$.
Each node in the CRF represents an image $I \in \mathcal{T} \cup \mathcal{W}$, and the label space is the quantized pose space sampled uniformly from yaw and pitch angles (360 poses  from $\text{yaw} \in [0^\circ, 360^\circ]$ and $\text{pitch} \in [0^\circ, 45^\circ]$, roll angle is fixed). 
For image $I$ we search for the $K$ nearest neighbors among a set of rendered images $\mathcal{R}$ in a 3D database.

The unary potential is defined by the number of nearest neighbors in the rendered image set with the same discretized pose:
  \begin{equation}
    E_{\text{{unary}}}^{(i)}=\exp \left[ -\sum_{\{ I_k | I_k \in \mathcal{N}_i^{(K)} \subseteq \mathcal{R} \}}\mathds{1}_{\theta_{i}=\theta_{k}} \right]
    \end{equation}
where $\mathds{1}$ is the indicator function and $\mathcal{N}_i^{(K)}$ denotes the set of the $K=6$ nearest neighboring images of $I_i$ in terms of HOG-distance.

\begin{figure}[t]
\begin{center}
 \includegraphics[width=0.99\linewidth]{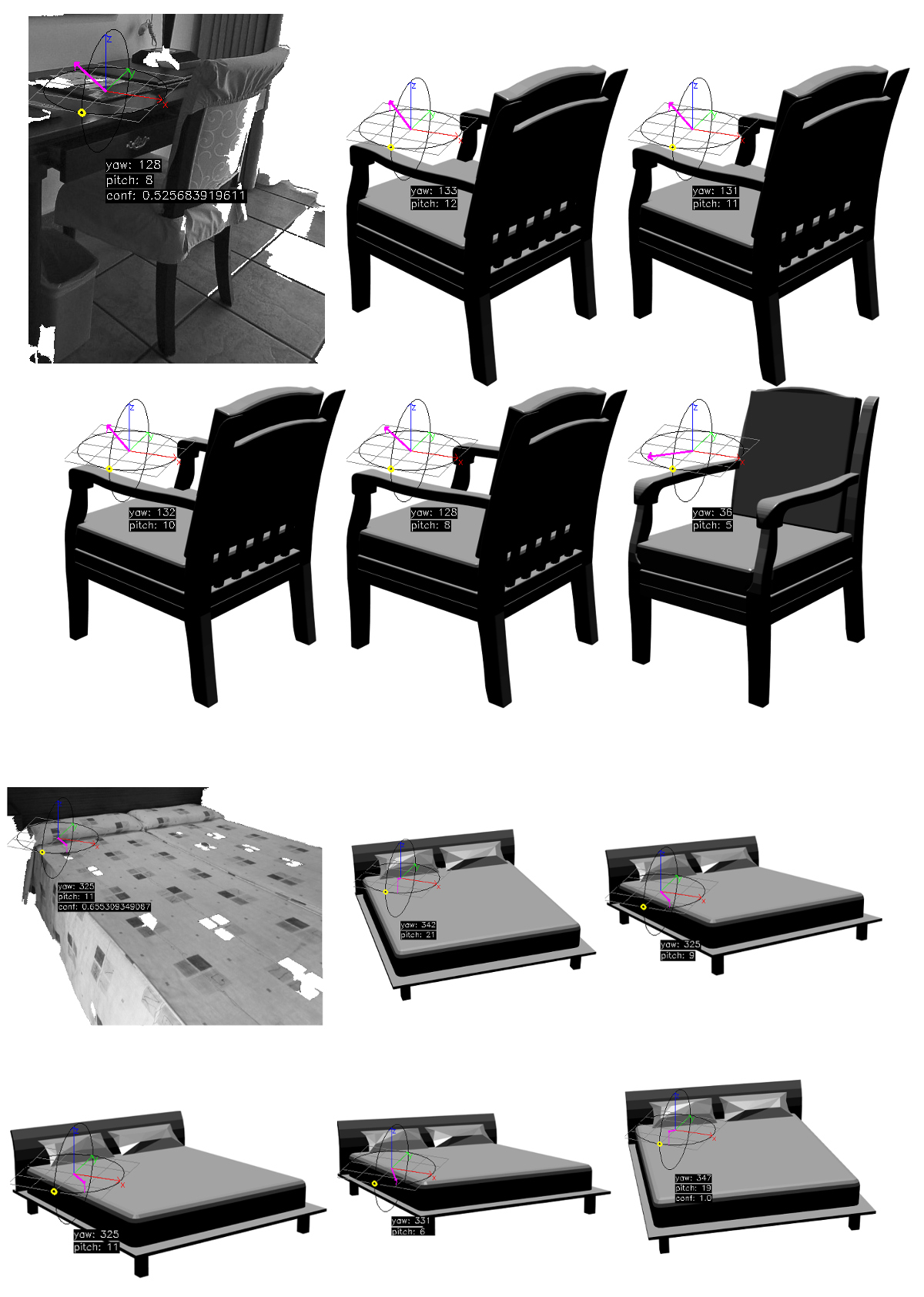}
\end{center}
\caption{\textbf{3D pose estimation:} \it Two pose estimation results for a segmented input image (top left) shown with the five 3D models closest in HOG space.
}
\label{figChair2Pose}
\end{figure}

The binary potential between two images $I_{i}$ and $I_{j}$ in $\mathcal{T} \cup \mathcal{W}$ encourages smoothness between the predicted poses of neighboring images:
  \begin{equation}
    E_{\text{\text{binary}}}^{(i,j)}= \, d^{\gamma}(\theta_{i},\theta_{j}) \; d^{HOG}(I_{i},I_{j}) ,
  \end{equation}
  where $d^{\gamma}(\theta_{i},\theta_{j})$ is the 
  an angle distance function defined as
\begin{align}
&  d^{\gamma}(\theta_{i}, \theta_{j})=\text{{min}}(d(\theta_{i}, \theta_{j}), \gamma ) \; , \textrm{where} \\
& d({\theta_{i}},{\theta_{j}}) = |\rho_i - \rho_j| + |\xi_i - \xi_j| ,
\end{align}
and $\gamma$ is a threshold, $\rho$ is the yaw angle, and $\xi$ the pitch angle. The energy function for the CRF is then:
\begin{equation}
  E_{CRF} = \sum_{I_i \in \mathcal{T} \cup \mathcal{W}} E_{\text{unary}}^{(i)} +
  \sum_{\{I_i \sim I_j | I_i, I_j \in \mathcal{T} \cup \mathcal{W} \}} E_{\text{binary}}^{(i,j)},
\end{equation}
and CRF inference is performed using the TRW-S algorithm~\cite{kolmogorov2006convergent}.
Qualitative results can be seen in Fig.~\ref{figChair2Pose}. 3D model retrieval is performed by retrieving the nearest neighbor in HOG space among the images in $\mathcal{R}$ and the cost function for $E_o$ for object orientation is the Euclidean distance of the descriptors.\\

\subsection{Context prior}


The context prior, $\pi(\Phi)$, evaluates the relative positions and orientations of objects and walls in a 2D top-down view of the scene.
The object-to-wall cost $\costterm_{o,w}$  measures distance and alignment of an object with its closest wall segment:
\begin{equation}
\label{eqn:Eow}
\costterm_{o,w} (\cparams) = \sum_{j=1}^{N_o} \Vert p^o_j - p^w_{i^*(j)} \Vert + \nu_{n} \sum_{j=1}^{N_o} \Vert {n^o_j}^\top  n^w_{i^*(j)} \Vert,
\end{equation}
where $p^o_j$ is the position of object $o_j$, $i^*(j) = \argmin_i d(p^o_j, p^w_i)$ is the index of the closest wall segment to object $o_j$, $n^o_j$ and $n^w_{i^*(j)}$ are the normals of the object and its closest wall, respectively, and $\nu_{n}$ is a weighting factor.
The object-to-object cost $\costterm_{o,o}$  is a function penalizing the overlap between objects.
\begin{equation}
\costterm_{o,o} (\cparams) = \sum_{j,k=1}^{N_o} A \left(b(o_j) \cap b(o_k)\right),
\end{equation}
where $A$ is the area of intersection between two object bounding boxes, denoted as $b$. The prior term combines the object-to-wall and object-to-object costs and is defined as
\begin{equation}
\label{eqn:contextprior}
\contextfunc(\cparams) = \exp \left[- (\costterm_{o,w}(\cparams) + \mu \costterm_{o,o}(\cparams))\right],
\end{equation}
where $\mu$ is a weighting factor.


\subsection{MAP estimation}



We use a sampling strategy to find room layouts with a maximum posterior solution, as defined in Eq.~\ref{eqn:map}.
From an initial estimate of 3D room geometry and 3D object pose, we use the context prior term to sample locations and orientations of objects, as well as the global scale parameter~$\lambda$. Scale is sampled uniformly within a fixed interval, while object locations are sampled from a normal distribution that has large variance in the object-camera direction, accounting for distance ambiguity, and small variance normal to this direction, giving high confidence to the location predicted by the detector.
Object orientation is sampled from a normal distribution with a mean of the orientation found in section~\ref{sec:orientation}.
For each of the $N_S$ configuration samples we evaluate the likelihood terms (Eq.~\ref{eqn:lhood}) and context prior (Eq.~\ref{eqn:contextprior}) and output the hypothesis with the maximum posterior value. Implementation details are given in the next section.









\section{Results} \label{sec:experiments}

\begin{figure}[t!]
\begin{center}
\subfloat[]{\includegraphics[width=0.4\columnwidth]{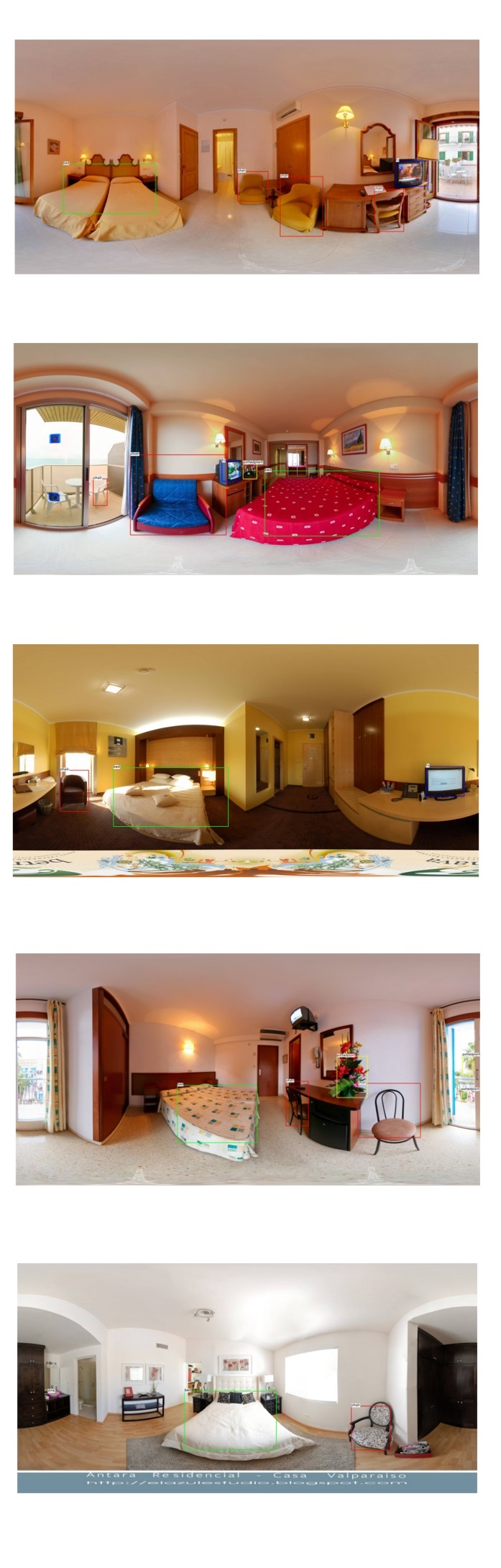}}
\subfloat[]{\includegraphics[width=0.285\columnwidth]{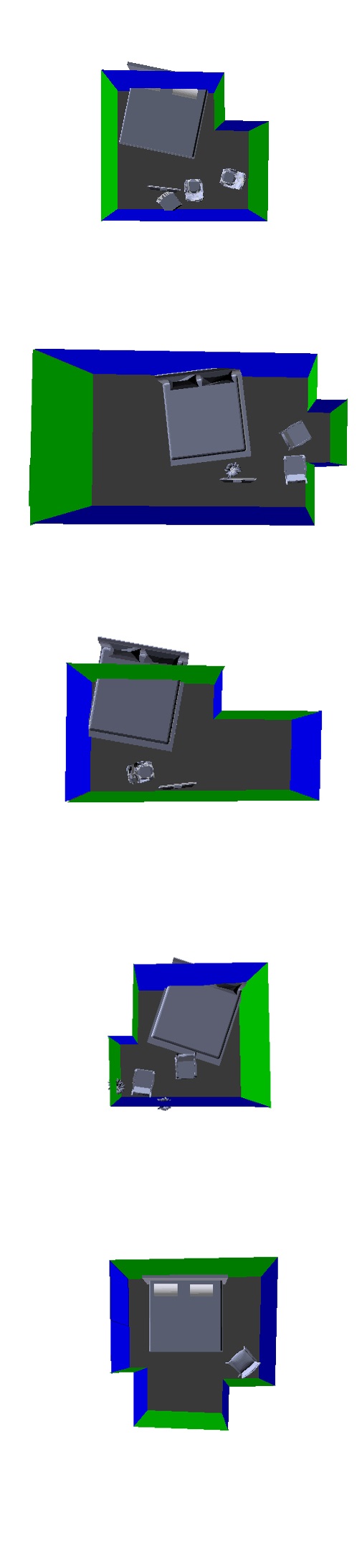}}
\subfloat[]{\includegraphics[width=0.315\columnwidth]{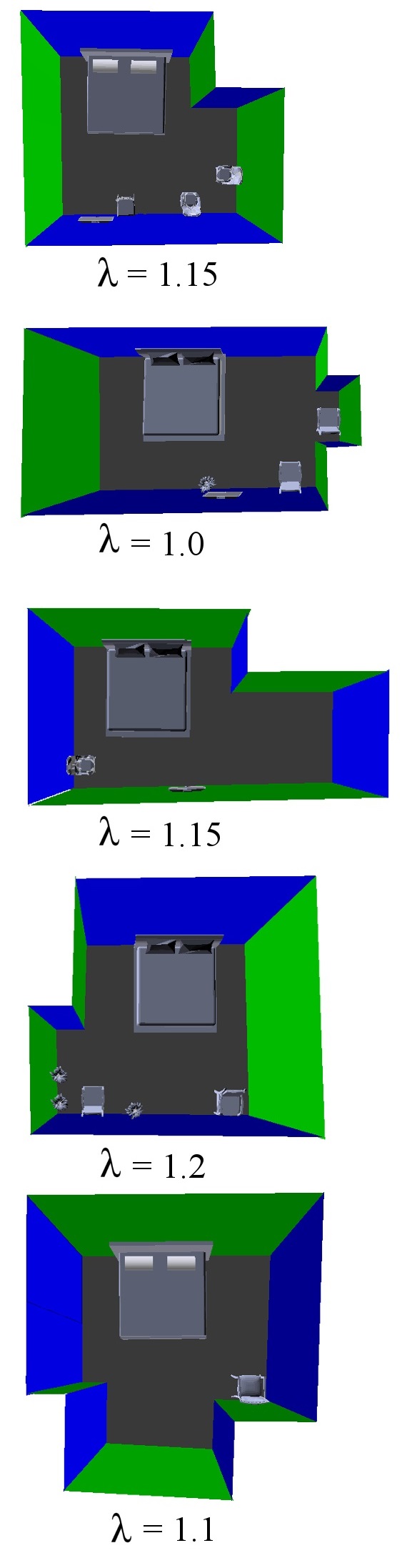}}\\
\end{center}
\caption{\textbf{Example results on SUN360 images:} \it (a) Panorama images with detected objects. (b) Initial layout from estimated surface orientations. (c) Optimized layout of our result. Input images from the SUN360 dataset are on the left. The center column shows the initial layout estimation from object detection and orientation surface (with unknown global scale), the right column shows results after use of the context prior.
Absolute wall height equals $2.5m \times \lambda$. The dimensions of 3D models from public datasets are known and remain fixed.}
\label{Fig5_Layout}
\end{figure}

\begin{figure}[ht!]
\begin{center}
 \includegraphics[width=\columnwidth]{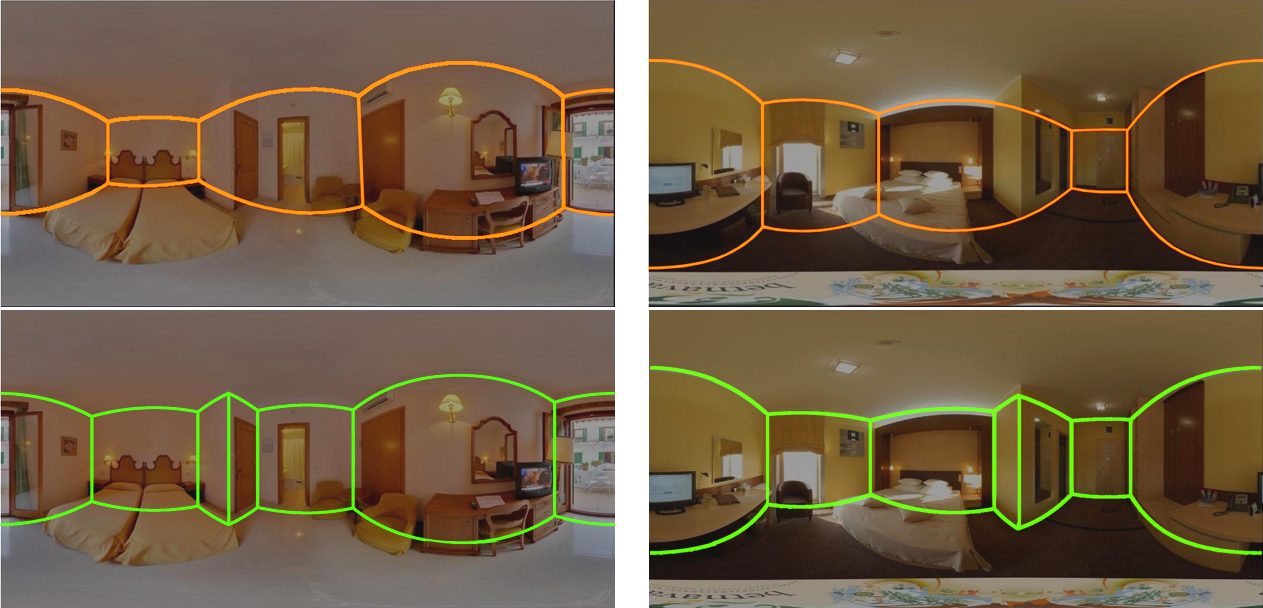}
\end{center}
\caption{\textbf{Room geometry comparison with {\it PanoContext}~\cite{ZhangECCV2014}.} \it (top) Two examples from the SUN360 dataset showing the limitation of the box geometry assumption. (bottom) Our method successfully recovers the room geometry in these cases. Images show the edges of walls projected into panorama images.}
\label{Fig_compare_panocontext}
\end{figure}

The algorithm was validated on a subset of the public SUN360 dataset~\cite{XiaoCVPR2012}.
The same dataset was used in prior work~\cite{ZhangECCV2014}.
The dataset contains panorama images of indoor scenes in high resolution (up to 9K) which are rescaled to 2K to reduce computation time.
We obtained reasonable initial pose estimations when object detection bounding boxes and segmentation were accurate (see Fig.~\ref{Fig5_Layout}(a)). However, Fig.~\ref{Fig5_Layout} (b) also shows that directly applying state-of-the-art techniques is insufficient to obtain correct room layouts.
Even though the alignment of surfaces estimated from different perspective views estimates the correct room shape, the absolute scale remains ambiguous.
In addition, the initial object pose estimation $\{\hat{\theta}^o_j\}$ (\eg, bed orientation) is not always accurate, and object distances to camera are unknown.
Our proposed method estimates more accurate results as shown in Fig. ~\ref{Fig5_Layout}(c).
A comparison of room geometry esitmation with the {\it PanoContext}~\cite{ZhangECCV2014} method is shown in Fig.~\ref{Fig_compare_panocontext}. We use code provided by the authors to generate 200,000 hypotheses per room as in~\cite{ZhangECCV2014}. These are ranked by surface normal consistency with the input panorama. The overall computation time is 11 minutes on an i7 processor. The top-ranked hypotheses are displayed in Fig.~\ref{Fig_compare_panocontext}, top row. 

\subsection{Quantitative evaluation}
\begin{figure}[t!]
\begin{center}
\subfloat[]{\includegraphics[width=0.5\columnwidth]{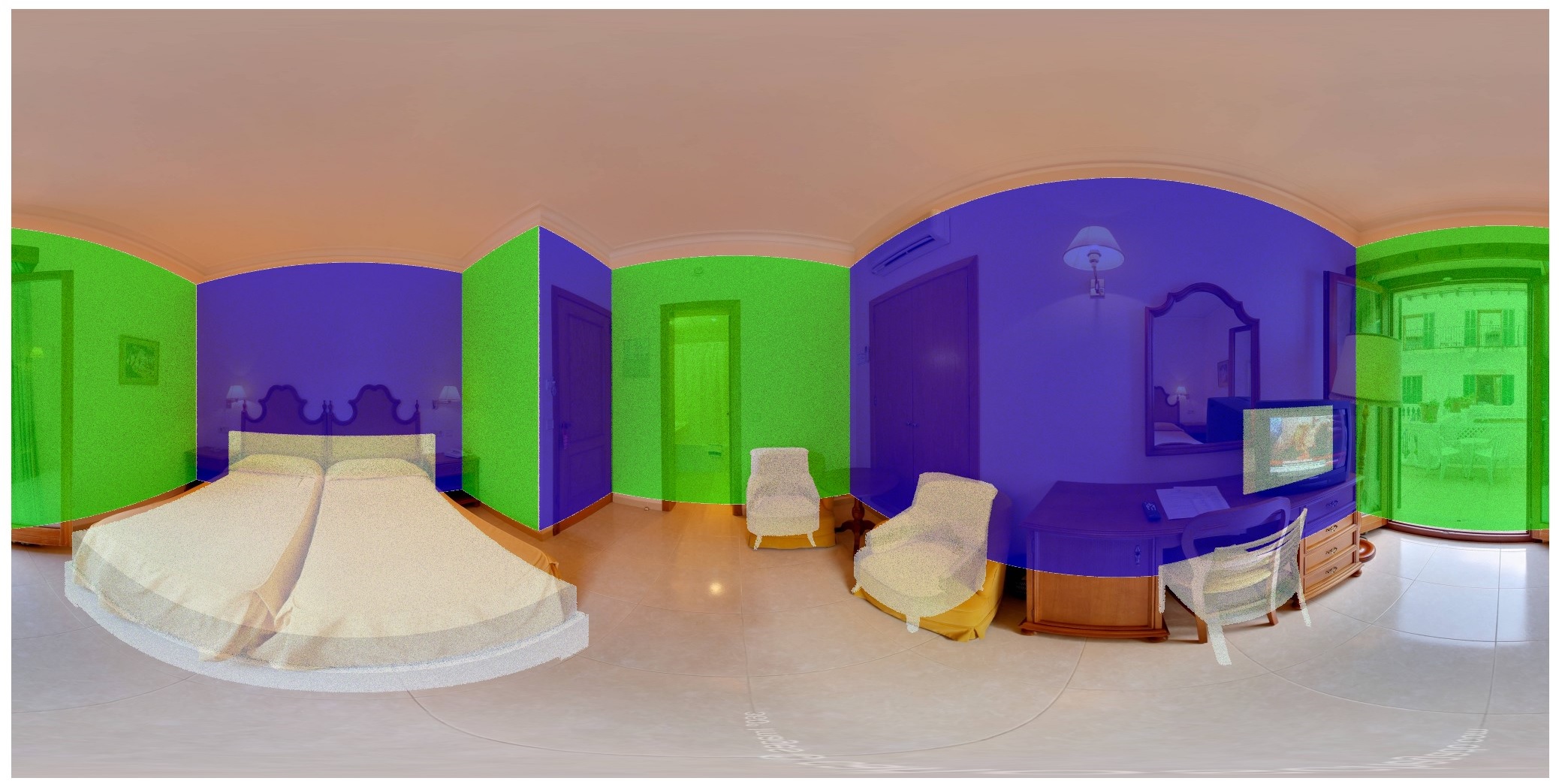}}
\subfloat[]{\includegraphics[width=0.5\columnwidth]{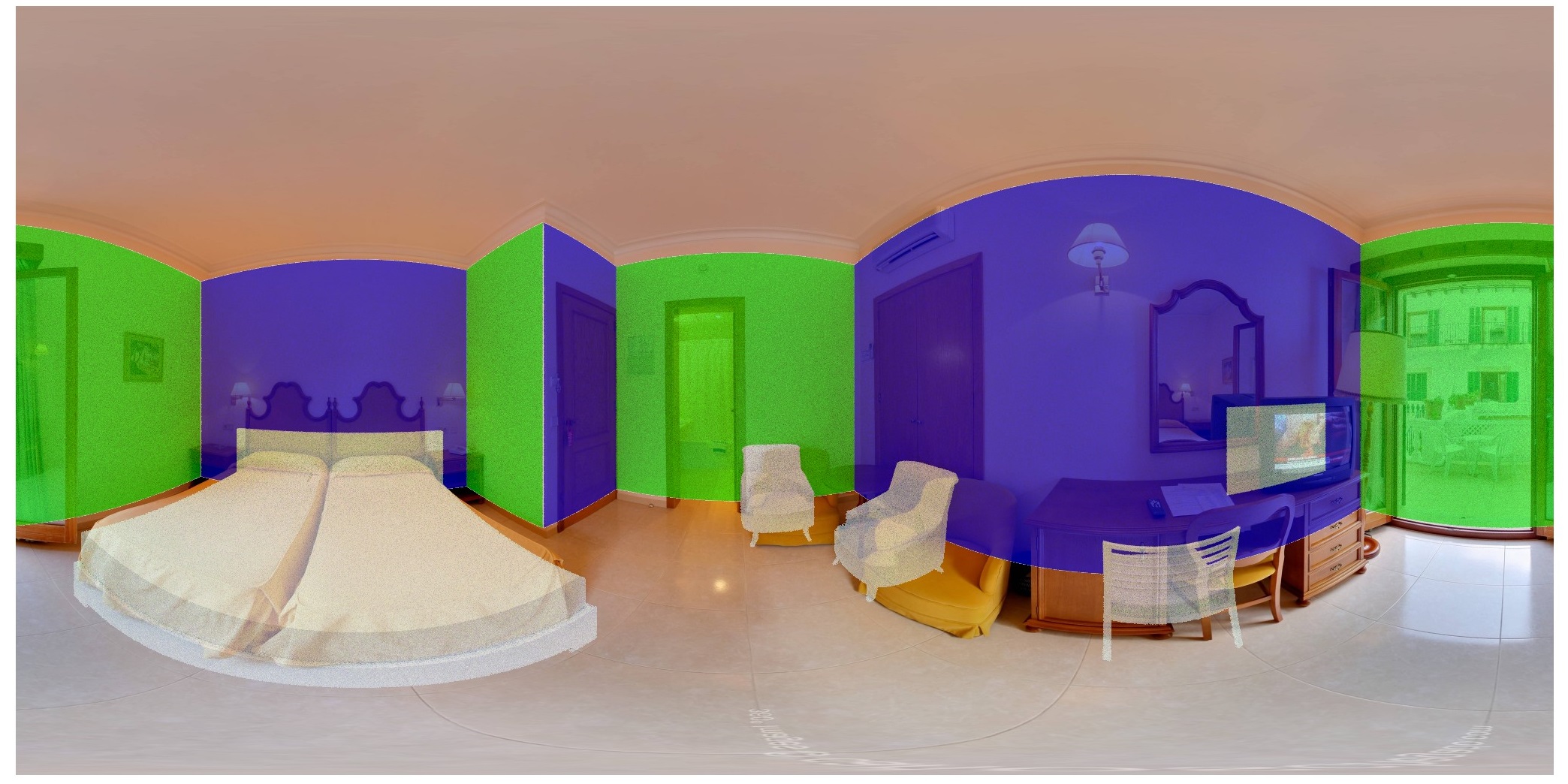}}\\
\end{center}
\caption{\textbf{Reprojection of 3D scene into panorama image:} \it  (a) Ground truth.  (b) Our result. Comparison of the estimated layout to manually annotated ground truth. Surface orientation and 3D objects are overlaid onto the input image.  Camera parameter approximations and shape differences between real objects and 3D models can cause slight misalignment.}
\label{Fig6_GT}
\end{figure}

To evaluate the accuracy of the proposed method, we manually annotated object positions and orientations in the panorama input images.
34 bedroom images are selected from SUN360 dataset and the results are shown in Table~\ref{tab:gteval}. 
We measure positional error as distance between object centroids projected onto the 2D ground plane and orientation error as the angle between ground truth and estimated pose.
As seen in Fig.~\ref{Fig6_GT}, the estimation error is lower for certain object classes, \eg, TV, where the pose can typically be estimated reliably and the object prior helps by favoring alignment with nearby walls.
The error for chairs tends to be higher as there is a large variation and symmetry of chair shapes and 3D pose estimation is less accurate.
Our method does not attempt to estimate the orientation of potted plants since they tend to be rotationally symmetric.
The joint estimation of room layout, scale and object pose also allows us to automatically generate a 2D floor map from one panorama image (see Fig.~\ref{Fig7_floormap}).

\begin{table}[t]
\setlength{\tabcolsep}{5pt}  
\begin{center}
\begin{tabular}{ l  c  c  c  c }
\toprule
    {\bf Object}    &  {\bf Position error (cm)}   &  {\bf Orientation error (deg)} \\ \midrule
Bed    & 25.0 $\pm$ 17.4 & 1.0 $\pm$ 1.4 \\ 
TV      & 4.7 $\pm$ 6.4 & 1.4 $\pm$ 1.1  \\ 
Chair & 52.3 $\pm$ 66.0 &  10.7 $\pm$ 15.0 \\ 
Plant & 8.7 $\pm$ 12.0 & - \\
\bottomrule
\end{tabular}
\end{center}
\vspace*{-1em}
\caption{\textbf{Evaluation on SUN360 images:} \it Object position and orientation errors measured against manually annotated ground truth.}
\label{tab:gteval}
\end{table}

\begin{figure}[t!]
\begin{center}
\includegraphics[width=\columnwidth]{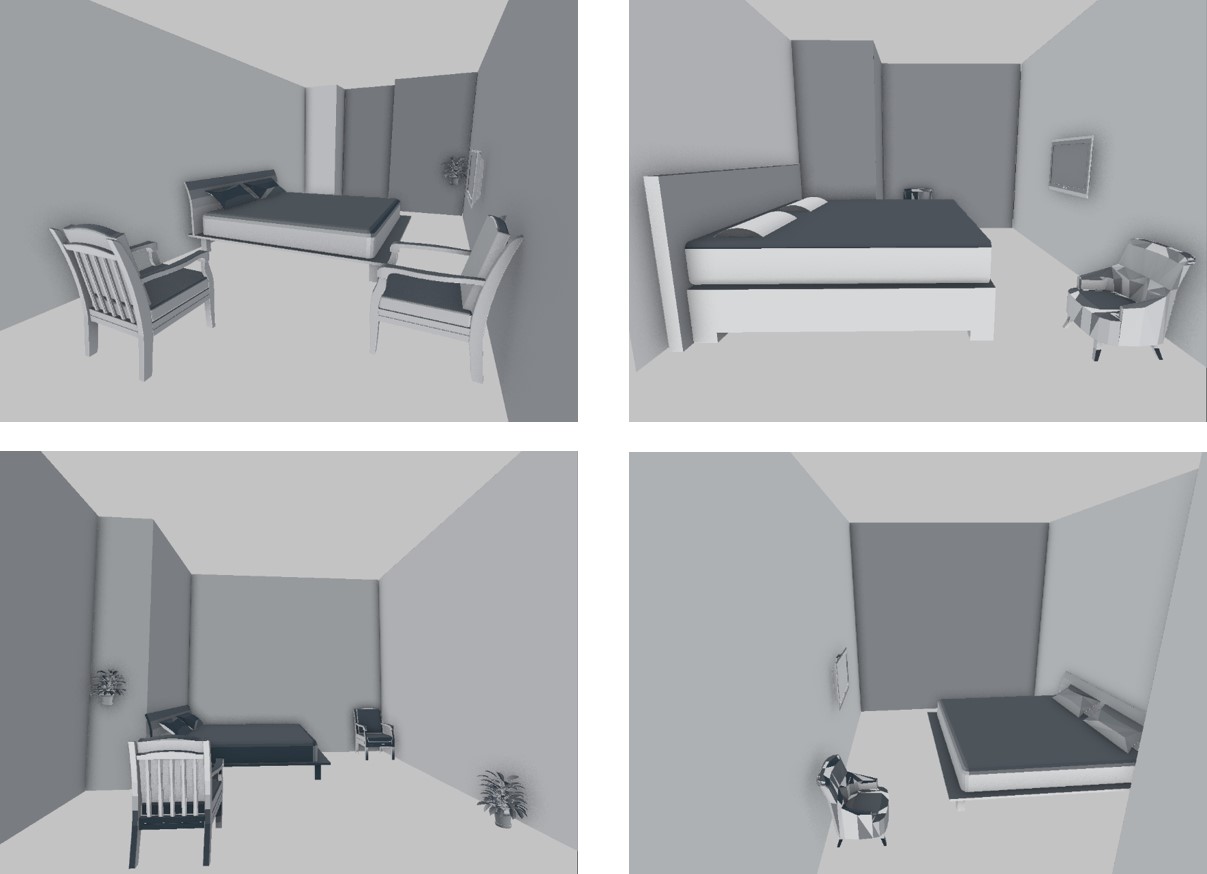}
\end{center}
\caption{\textbf{Synthetic room data.} \it Four example rooms out of 88 that were used in our quantitative evaluation.}
\label{fig:sytheticdata}
\end{figure}

To further assess the method's accuracy, including scale estimation, we evaluate it by generating 3D scenes as ground truth. We synthesize 88 rooms of arbitrary shape, based on existing room templates, and size, containing objects. See Fig.~\ref{fig:sytheticdata} for examples. Wall heights are sampled from a normal distribution with mean 2.7m and 0.2m standard devation. We add an offset to the length of each wall (only if longer than 0.7m), uniformly sampled offset from [-0.3m, 0.3m]. Objects are placed at random, and their position and orientation are updated by sampling from the context prior. Experimental results are reported in Table~\ref{tab:quant} and show the contributions of each step separately (object pose, room scale/wall height estimation).
Overall estimation using $N_S=3000$ samples is accurate to 0.8-8$^\circ$ and 2-20cm, depending on object class.

\begin{table*}[t!]
\renewcommand\thetable{2}
\centering
\begin{tabular}{ p{2.2cm} cccc p{0.2cm} cccc} 
\toprule
\small{{\bf Average errors}}   & \multicolumn{4}{c}{\small{{\bf after initialization}}} && \multicolumn{4}{c}{\small{{\bf with context term}}} \\
 &  \  \  \small{bed} &  \  \  \small{chair \  \ } &  \  \  \small{TV}& \small{plant} && \small{bed} & \small{chair} & \small{TV} & \small{plant}\\ \midrule
$\epsilon_\textrm{obj. orient.}$\small{(deg)}  &  \  \  \small{5.2$\pm$0.5}  \  \   &  \  \   \small{4.1$\pm$1.8}  \  \  &  \small{2.8$\pm$1.5}  & \small{n/a}  &&
\small{0.8$\pm$0.6} &  \small{8.0$\pm$6.4}  \  &  \small{0.8$\pm$0.7}  &  \small{n/a}  \\ \cmidrule{2-5} \cmidrule{7-10}
$\epsilon_\textrm{obj. pos.}$\small{(cm)} &  \small{197.6$\pm$57.6} & \small{186.7$\pm$99.6} & \small{156.21$\pm$73.0} & \small{174.7$\pm$70.3}   &&
\small{21.0$\pm$13.0} & \small{7.1$\pm$7.3} & \small{2.0$\pm$7.0} & \small{6.9$\pm$0.5} \\ \cmidrule{2-5} \cmidrule{7-10}
$\epsilon_\textrm{wall height}$\small{(cm)}    & \multicolumn{4}{c}{\small{n/a (initialized at 2.5m)}} &&  \multicolumn{4}{c}{\small{4.9$\pm$0.1}} \\
\bottomrule 
\end{tabular}
\vspace*{0.5em}
\caption{\textbf{Evaluation on synthetic dataset of 88 rooms.} \it The table shows the mean error with standard deviation of object orientations, object positions, and wall height. The benefit of the proposed context prior is shown by comparing the results after the initialization stage (left) and after including context-based sampling (right). Average chair orientation error increases slightly. Note that the orientation error of potted plants is omitted, since they do not have a canonical orientation.}
\label{tab:quant}
\end{table*}

\begin{figure*}[ht!]
\begin{center}
 \includegraphics[width=\linewidth]{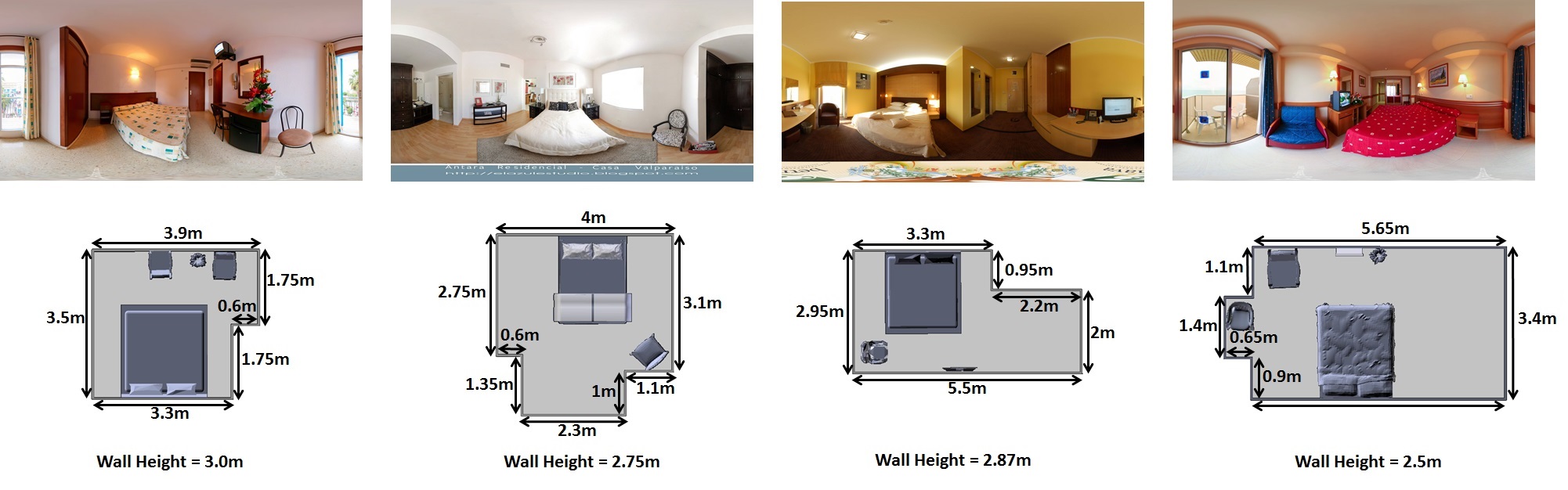}
\end{center}
\caption{\textbf{Automatic 2D floor map generation with estimated scale.} \it The method estimates the global scale by transferring the scale of known 3D objects to scene objects. Examples shown were generated from SUN360 images.}
\label{Fig7_floormap}
\vspace*{-2mm}
\end{figure*}

\begin{figure}[ht!]
\begin{center}
 \includegraphics[width=\linewidth]{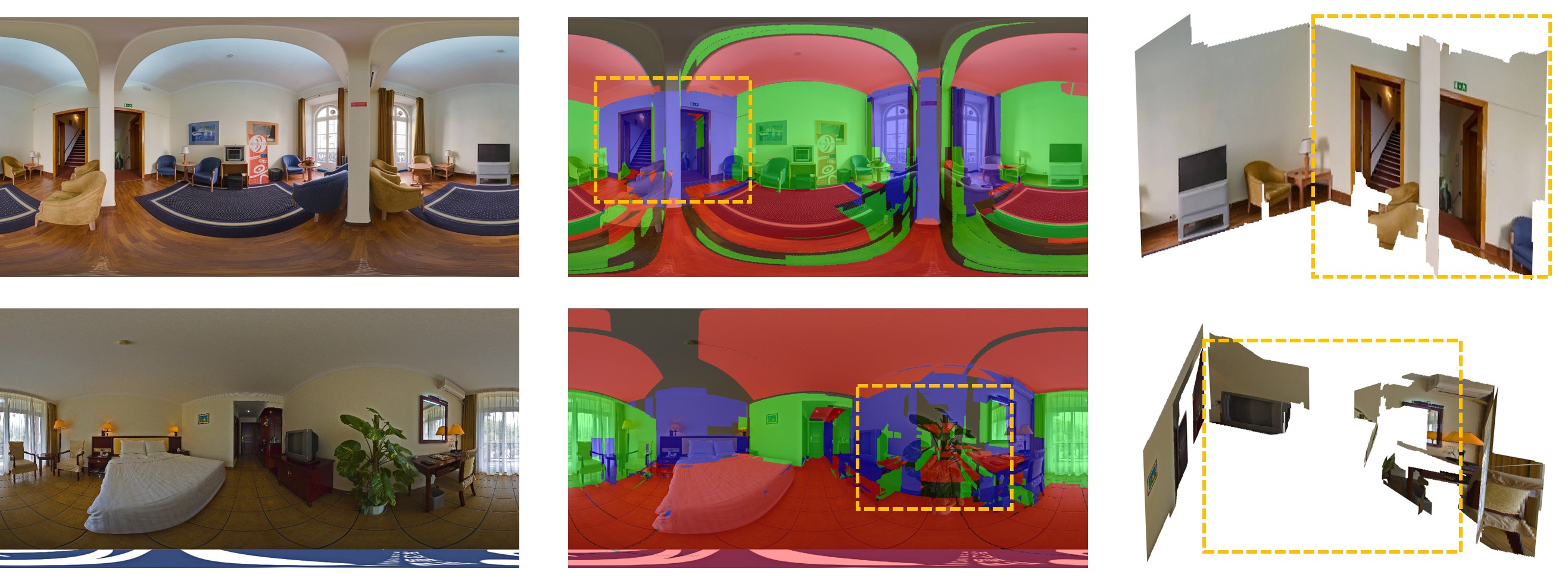}
\end{center}
\caption{\textbf{Examples of failure cases.} \it The surface orientation estimation currently fails on thin stuctures (top) and irregular shapes like large plants (bottom), leading to incorrect estimation of room geometry.}
\label{Fig_failurecases}
\end{figure}

Note that failure cases can stem both from failed object detection and from incorrect surface estimation. Two typical failure cases of surface orientation recovery for thin and irregular structures are shown in Fig.~\ref{Fig_failurecases}.

\subsection{Implementation details}
\noindent \textbf{Object detection. }
We use the Faster R-CNN from~\cite{ShaoqingRenNIPS2015} for object detection and recognition.
The MS COCO~\cite{LinArxiv2014} dataset is used for training, since it contains a large number of indoor object categories (\eg,  {\it chair, couch, potted plant, bed, dining table, toilet, tv, laptop, microwave, oven, refrigerator, clock, vase}).
The model was trained on the 80,000 image training set for 240,000 iterations with VGG16 networks~\cite{SimonyanArxiv2014}, using top-2000-score Region Proposal Networks (RPN)~\cite{ShaoqingRenNIPS2015} and  Multi-scale Combinatorial Grouping (MCG)~\cite{PontTusetArxiv2015} as object proposals. The mean average precision (mAP) for all 80 object classes is 49.0\%,  and 26.5\% for the intersection over union (IoU) values of $50\%$ and $95\%$, respectively. The same metric and validation set was used for the MS COCO 2014 primary challenge. Note that our detection performance is competitive with the current state-of-the-art method by He~\etal~\cite{HeCVPR2016},  which achieved corresponding mAP scores of $48.4\%$ and $27.2\%$.\\

\noindent \textbf{Object pose estimation. }
We collected a set of 3D models from the 3D Warehouse~\cite{warehouse2015} and rendered each in 360 poses.
For hotel rooms, we used 9 beds, 16 chairs, 4 plants, 6 TVs, and crawled 300-350 Internet images per class using Google image search.
Note that results in Table~\ref{tab:gteval} are for detected objects only.
The number of nearest neighbors $K$ is set to $6$.
The truncation threshold $\gamma$ is set to $20^{\circ}$. TRW-S is run for $100$ iterations to estimate object pose.\\

\noindent \textbf{Context prior and sampling. }
Scale, object location and orientation are sampled by evaluating the context prior. We use 8 sampling epochs of 25 samples each. In every epoch the sample with largest context prior term is used as seed sample for the next epoch. The normal sampling distribution along the camera-to-object-center direction has the obtain location as mean, and a variance of 0.1 times the camera-object distance, and a variance of 0.005 times this distance along the perpendicular direction. Orientation is sampled from a normal with variance $0.1$ rad, and scale is sampled uniformly, in terms of wall height, from the interval [2.0m, 3.5m]. The weight $\nu_{n}$ in Eq.~\ref{eqn:Eow} is set to 10.0, and $\mu$ in Eq.~\ref{eqn:contextprior} to 0.25.\\

\noindent \textbf{Computation time. } The layout estimation pipeline was implemented on a desktop PC with i7 processor and 8GB RAM. The main bottleneck is currently the object pose estimation step using CRF optimization, which takes approximately 1-2min per object class. The object detection method in the pipeline takes 7s on average for 18 perspective images using a GRID K520 GPU. One room layout hypothesis evaluation requires about 30s.



\section{Conclusions}

In this paper we presented a new formulation for indoor layout estimation.
We demonstrated its ability to recover complex room shape with Manhattan World assumption from a single panorama image using detected objects, their pose and their context in the scene.
The proposed method does not rely on video, multiple images, or depth sensors as input~\cite{CabralCVPR2014} and is not limited to simple box shaped rooms as in recent work on panoramic reconstructions~\cite{ZhangECCV2014}.
Compared to~\cite{YangCVPR2016}, our method takes the class, location, and pose of objects into account and introduce a context prior to this underconstrained problem.
We evaluated the method quantitatively on a synthetic dataset and qualitatively on images from the SUN360 dataset.
One limitation of the proposed method is that it currently relies on the output of an object detector. Objects that are not detected are currently not part of the final 3D model. 
Recent CNN-based methods for predicting depth and semantic labels~\cite{EigenICCV2015} or 3D object pose~\cite{SuICCV2015} from images may be leveraged to improve the results.


\bibliographystyle{ieee}
\small
\bibliography{layout}
\end{document}